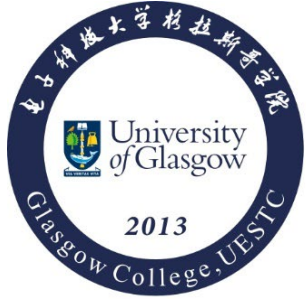

**Final Year Project Report**

**Bachelor of Engineering**

**Research on Robot Path Planning Based on Visual SLAM**

Student: Wang Ruiqi

GUID: 2614425    UESTC ID: 2020190506022

Supervisor:

2023-24

**Coursework Declaration and Feedback Form**

*The Student should complete and sign this part*

| Student Name: Wang Ruiqi | Student GUID: 2614425 |
|---|---|
| Course Code: UESTC4006P | Course Name: INDIVIDUAL PROJECT 4 |
| Name of Supervisor: | |

| Title of Project: Research on Robot Path Planning Based on Visual SLAM |
|---|

## Declaration of Originality and Submission Information

| *I affirm that this submission is all my own work in accordance with the University of Glasgow Regulations and the School of Engineering requirements*<br>Signed (Student): | 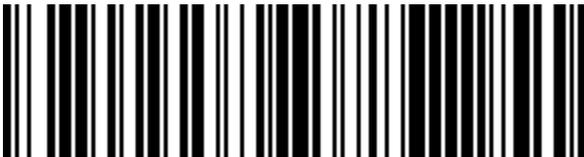 |
|---|---|

| Date of Submission: |
|---|

| *Feedback from Lecturer to Student – to be completed by Lecturer or Demonstrator* |
|---|

Grade Awarded:

Feedback (as appropriate to the coursework which was assessed):

| Lecturer/Demonstrator: | Date returned to the Teaching Office: |
|---|---|

# Abstract

With the rapid development of robotics technology, robots are playing an increasingly important role in people's daily lives. Autonomous navigation ability is the prerequisite for robots to complete other tasks independently. As the key technologies in autonomous navigation, SLAM and path planning have become the focus of research in the field of robotics in recent years.

This project has conduct research on robot path planning based on Visual SLAM. The main work of this project is as follows:

Construction of Visual SLAM system. Research has been conducted on the basic architecture of Visual SLAM. A Visual SLAM system is developed based on ORB-SLAM3 system, which can conduct dense point cloud mapping.

The map suitable for two-dimensional path planning is obtained through map conversion. This part converts the dense point cloud map obtained by Visual SLAM system into an octomap and then performs projection transformation to the grid map. The map conversion converts the dense point cloud map containing a large amount of redundant map information into an extremely lightweight grid map suitable for path planning.

Research on path planning algorithm based on reinforcement learning. This project has conducted experimental comparisons between the Q-learning algorithm, the DQN algorithm, and the SARSA algorithm, and found that DQN is the algorithm with the fastest convergence and best performance in high-dimensional complex environments.

This project has conducted experimental verification of the Visual SLAM system in a simulation environment. The experimental results obtained based on open-source dataset and self-made dataset prove the feasibility and effectiveness of the designed Visual SLAM system. At the same time, this project has also conducted comparative experiments on the three reinforcement learning algorithms under the same experimental condition to obtain the optimal algorithm under the experimental condition.

**Keywords:** Visual SLAM, map conversion, path planning, reinforcement learning

# Acknowledgements

In the process of completing my individual project, I would like to express my deepest gratitude to many people.

First, I would like to thank my family for their understanding and support throughout my undergraduate studies. Your companionship and encouragement have enabled me to overcome difficulties and persevere in pursuing my goals.

At the same time, I would also like to thank my classmates and teachers for giving me a happy and fulfilling undergraduate life. The memories of this period will become my spiritual treasure years later.

I would also like to express my sincere gratitude to my instructor, Professor Lin Jie. Thank you for guiding me and answering my questions during the research process. With your help, I finally complete my individual project successfully.

In addition, I would also like to thank Teacher Liu Lei from the School of Aeronautics and Astronautics. Thank you for providing me with experimental equipment when I was short of it. It was only with your generous help that I was able to successfully complete the experiment.

Finally, I would like to thank my hard work during my undergraduate studies, which enables me to continue my next stage study. All endings are new beginnings. Walk on, see you on the top!

# Contents

# 1 Introduction

## 1.1 Research Background and Meaning

With the vigorous development of robotics technology, robots have been widely used in industrial production, military and national defence, and daily life. In some environments, especially unknown or extreme environments, robots can replace humans to complete work while ensuring the safety of people's lives and property.

However, to make robots complete tasks and meet various work needs, it is a prerequisite for robots to be able to perceive and interact with the external environment like humans. Therefore, research on map construction and path planning of mobile robots is necessary. In the actual environment, the mobile robot obtains information about the surrounding environment through its sensors and constructs a map. The constructed map can be used to plan the moving path and realize the navigation of the mobile robot while avoiding collisions with obstacles. The process of robot movement in the unknown environment can be divided into two parts:

(1) Map construction: Robot map construction refers to the process in which the robot uses various sensors (such as lidar, cameras, ultrasonic sensors, etc.) to obtain information about the surrounding environment, and then uses this information to build a map. The map can be two-dimensional or three-dimensional and is used to represent the environment where the robot is located, including obstacles, roads, and other information. SLAM (Simultaneous Localization and Mapping) technology allows the mobile robot to build an environment model while moving in a previously unknown environment and estimate its movement at the same time.

(2) Path planning: After obtaining the robot's position and environmental information through SLAM, the path-planning algorithm is used to obtain an optimal path from the robot's position to the target position. An excellent path-planning algorithm can find the shortest path from the starting point to the target point for the mobile robot while avoiding collisions with obstacles.

Currently, there are many excellent research results in the two independent research areas of map construction and path planning. However, there is a relative lack of relevant research on a complete system that combines the two. In addition, the high cost (laser) and low accuracy (inertial, radar) of some SLAM sensors make SLAM systems less practical in real life. Considering the above problems, this project takes robot path planning based on Visual SLAM as the research topic, aiming to study how to use visual sensors to realize robot localization and map construction and obtain the optimal path for moving robots.

## 1.2 Research Status

### 1.2.1 Visual SLAM

R.C. Smith and P. Cheesement's research on the representation and estimation of spatial uncertainty in 1986 is the early pioneering work in SLAM [1]. Subsequently, in 1995, Durrant-Whyte combined the map creation and localization problems and formulated them as an estimating problem, the concept of SLAM was first proposed [2]. After that, more and more scholars participated in SLAM research, and the SLAM field developed vigorously. Depending on the type of sensor, SLAM has derived many branches, including Visual SLAM, Inertia SLAM, Depth SLAM, etc., in which Visual SLAM is the most widely used due to its low cost and high accuracy.

MonoSLAM proposed by A. J. Davison in 2007 [3] is a landmark work in Visual SLAM and is the first real-time monocular visual SLAM system. Before MonoSLAM, Visual SLAM could not run online and could only rely on robots to collect data and then perform localization and mapping offline. MonoSLAM uses the extended Kalman filter (EMF) as the backend to track very sparse feature points at the front-end and can create continuous sparse maps online. However, sparse feature points are particularly prone to loss, which prevents future development of MonoSLAM. Klein et al. proposed the PTAM (Parallel Tracking and Mapping) [4] in 2007. PTAM proposed some new ideas for Visual SLAM: it proposed the parallelization of the tracking and mapping process and distinguished the concepts of front-end and back-end for the first time, which points out the direction for the Visual SLAM system that appears later. In addition, PTAM is the first scheme to use nonlinear optimization in the backend instead of using EKF. It introduces a keyframe mechanism. The system no longer needs to process each frame of image, but focuses on the extracted key images, and then optimizes the trajectory and map, which reduces the computational burden. PTAM is a milestone achievement in the history of Visual SLAM development. However, like MonoSLAM, PTAM also has the disadvantages of small application scenarios and easy tracking loss. These shortcomings have been prioritized in the future Visual SLAM systems. One of the most famous successors of PTAM is ORB-SLAM. Raul Mur-Artal proposed the first-generation ORB-SLAM algorithm in 2015 [5], which innovatively divides the system into three independent threads: tracking, local mapping and loop closing. ORB-SLAM is calculated around ORB feature points, including visual odometry and the ORB dictionary. In 2017, ORB-SLAM2 [6] was proposed as the improved of ORB-SLAM. The algorithm includes monocular, binocular and RGB-D camera modes, which improves localization accuracy while ensuring real-time performance. ORB-SLAM3 raised in 2020 [7] further improves the robustness, localization accuracy, loop detection and other aspects of ORB-SLAM2, making it perform better in complex environments and have a wider range of application scenarios.

### 1.2.2　　　　Path Planning

Path planning is the key part in robot autonomous navigation. Its goal is to enable mobile robots to independently find the shortest collision-free path from the starting point to the target point based on the map information obtained by the sensor. According to the planning scope and application scenarios, path-planning algorithms can be divided into local path planning (LPP) and global path planning (GPP). LPP is the process of planning a collision-free path for the robot from its current location. The goal is to ensure that the robot is safe in the local environment and reaches the target location quickly. GPP refers to path planning in the entire map and environment to find the shortest collision-free path from the starting point to the target point. LPP is usually used as a supplement to GPP so that the robot can adjust its path to adapt to changes in the environment during movement. Path planning algorithms are usually classified into three groups: search-based path planning, sampling-based path planning, and intelligent algorithm-based path planning.

#### 1.2.2.1　　　　Search-Based Path Planning

Dutch computer scientist Edsger W. Dijkstra proposed Dijkstra's algorithm in 1959 to find optimal paths [8]. The algorithm iterates from the starting point and gradually expands, finding the shortest path for a node at each step. It is the first widely used path-planning algorithm, but the computational complexity is high for real-time path planning and large-scale maps. In 1968, Peter E. Hart proposed the A* algorithm based on the Dijkstra algorithm [9], introducing heuristic functions, cost functions and priority queues. These improvements enable the A* algorithm to find the shortest path faster. The high efficiency of the A* algorithm has made it widely used, and its creative invention of introducing heuristic functions has also been recognized by many researchers. To meet the needs of different application scenarios, many derivative algorithms based on the A* algorithm have also emerged. The Anytime Repairing A* algorithm [10] is an incremental path planning algorithm based on the A* algorithm, which can correct the path in real-time in a dynamic environment to adapt to environmental changes. The D* algorithm [11] is also an incremental path planning algorithm used to quickly update paths when unknown map changes occur. There are many similar derived algorithms, including

Real-Time Adaptive A* [12], Learning Real-Time A* [13], D* Lite [14], Focussed D* [15] and so on. The development of these algorithms reflects the continuous efforts to optimize search-based path planning algorithms to adapt to maps of different scales, dynamic environments, and real-time requirements.

**1.2.2.2    Sampling-Based Path Planning**

The RRT (Rapidly-exploring Random Tree) algorithm proposed by Lavalle et al. in 1998 [16] is a pioneer of sampling-based path-planning algorithms. The RRT algorithm gradually builds a tree through random sampling and rapid exploration to effectively search and discover feasible paths. This algorithm only needs to consider the collision and superposition of random sampling points and map space when performing path search and does not need to know the state of the entire map space, which is beneficial to solving path search problems in high-dimensional and complex spaces. However, the RRT algorithm's efficiency is modest and has risks in falling into the problem of locally optimal solutions, resulting in the final path obtained is not always the optimal path. Kuffer et al. proposed RRT-connect algorithm based on the RRT algorithm [17]. The RRT algorithm is a one-way expansion from the starting point to the target point, while the RRT-connect algorithm expands two trees at the same time, one starting from the starting point and the other starting from the target point until the two trees meet. This improves the convergence speed of the algorithm. Sertac et al. proposed the RRT* algorithm [18] to solve the problem that the RRT algorithm does not always obtain the optimal path. The RRT* algorithm introduces a reconnection strategy that can optimize existing paths during the expansion of the tree. As the number of samples increases, the RRT* algorithm will progressively converge to the optimal path. Like the A* algorithm, the RRT algorithm also has many derivative algorithms. In addition to the previously mentioned RRT-connect and RRT*, there are also Extended-RRT [19], Dynamic-RRT [20], Anytime RRT* [21] and other algorithms. These algorithms enable sampling-based path-planning algorithms to meet the requirement of different application scenarios.

**1.2.2.3    Intelligent Algorithm-Based Path Planning**

Search-based and sampling-based path-planning algorithms are both traditional path-planning algorithms, and these algorithms can solve path-planning problems in most environments. However, in the face of complex dynamic environments, the problem of insufficient adaptability of traditional algorithms will be exposed. Fortunately, with the tremendous development of artificial intelligence in recent years, path-planning algorithms based on intelligent algorithms can successfully address the problem of poor flexibility of traditional algorithms in complicated dynamic contexts.

Early intelligent algorithm path-planning methods mainly focused on genetic algorithms [22] and simulated annealing algorithms [23]. These algorithms search for optimal paths by simulating the processes of biological evolution and material annealing, but their application in complex environments is subject to certain limitations. The ant colony algorithm [24] simulates the behaviour of ants releasing pheromones and finding paths during the process of searching for food and is used in path planning. Ant colony algorithms can search for better paths in complex environments, so it has been widely applied in the field of path planning.

With the rapid development of RL (Reinforcement Learning) in recent years, it has also been applied in the path planning problems of mobile robots. RL does not require the robot to have prior knowledge of the external environment. Instead, it learns during the continuous interaction between the robot and the environment, allowing the robot to avoid obstacles autonomously and complete path planning.

# 2 Research on relevant theories

## 2.1 Research on Visual SLAM

As seen in the diagram, the visual SLAM framework is primarily composed of five components shown in figure 1. The visual sensor gathers and preprocesses picture data from the surrounding environment. The VO (Visual Odometry) conducts local mapping and calculates the movement of the sensor based on the sensor's acquisition of neighbouring picture data. Whether the sensor returns to its initial position is detected by Loop Closure Detection. The back-end Optimisation part receives the camera posture and loop detection information calculated by the VO and Loop Closure Detection and splices the local maps to get the global trajectory and map. Finally, global mapping is completed based on the projected trajectory and map. A complete Visual SLAM system is a huge and complex system, which requires a large amount of relevant knowledge to fully understand it. The remainder of this section aims to give a brief introduction of three part of Visual SLAM: Visual Odometry, Optimization, and Loop Closure Detection, which may be unfamiliar to readers. Algorithm details and mathematical derivation are not included in this section. Interested readers please refer to references.

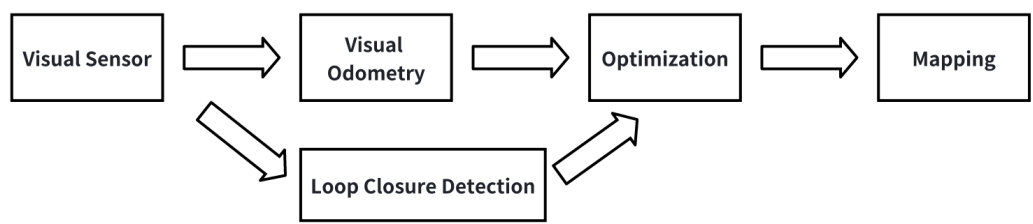

**Figure 1.** The framework of Visual SLAM.

### 2.1.1 Visual Odometry

Visual Odometry algorithms can be divided into two categories: Feature-based Method and Direct Method. In contrast to the Direct Method, the Feature-based Method has the advantages of stability and insensitivity to lighting conditions and moving objects. Therefore, it is the current mainstream solution. This section mainly introduces the Feature-based Method.

In the Feature-based Method, the Visual Odometry extracts feature points from the input frame and matches the extracted feature points to obtain matching feature point pairs between adjacent frames. The matching feature point pairs are used to estimate the movement trajectory of the sensor between adjacent frames.

#### 2.1.1.1 Feature extraction

Key points and descriptors make up feature points, which are unique places within the picture. The feature point's position in the picture is referred to as the key point. Information on its size and orientation may also be included. A vector that represents the data from the pixels surrounding the important point is the descriptor in most cases.

Researchers have proposed many feature point extraction algorithms, including SIFT, SURF and ORB. SIFT (Scale-Invariant Feature Transform) [25] takes into account the changes in illumination, scale, rotation, and so on that occur throughout the picture transformation process; yet it requires a significant amount of calculation work. Based on SIFT, SURF (Speeded Up Robust Features) [26] improves the calculation speed by using acceleration technologies such as integral images and scale space scaling. ORB (Oriented Fast and Rotated BRIEF) [27] uses FAST key points and BRIEF descriptors. Compared with the first two algorithms, it achieves a significant improvement in feature extraction speed and can meet real-time requirements. According to the experimental results [27], ORB has dominant advantage in feature point extraction speed over SIFT and SURF.

**Table 1.** Time consumption of extracting 1000 feature points with SIFT, SURF and ORB

| Feature point extraction algorithm | Number of extracted feature points | Time consumption |
|---|---|---|
| SIFT | 1000 | 5228.7 ms |
| SURF | 1000 | 217.3 ms |
| ORB | 1000 | 15.3 ms |

A type of corner point known as the FAST key point primarily finds areas where discernible local grayscale shifts occur. FAST is substantially faster than conventional corner detection methods since it simply compares pixel brightness. ORB includes scale and rotation descriptors to compensate for FAST corner points' lack of directionality and scale. Scale invariance is achieved by constructing an image pyramid shown in figure 2 and identifying corner points at each level. The grayscale centroid approach achieves directionality.

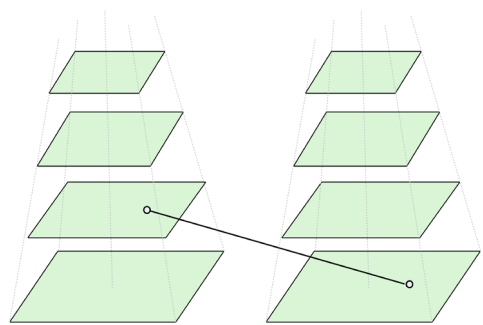

**Figure 2.** Corner points in different levels of image pyramid can be matched

BRIEF (Binary Robust Independent Elementary Feature) is a very fast binary descriptor with a description vector of 0 and 1. The values 0 and 1 describe the size connection between two random pixels (m and n) surrounding the key point: if m is greater than n, take 1, else take 0. ORB adds direction information to FAST feature points and gives the descriptor significant rotation invariance by computing the "Steer BRIEF" feature [27].

#### 2.1.1.2　　Feature matching

Feature matching is an extremely important step after feature point extraction. In Visual SLAM, feature matching solves the data association problem by determining the correlation between the item now viewed and the same object observed before. Accurate matching of feature points between images can reduce the burden on subsequent pose estimation and mapping.

Brute force matching is the simplest feature matching method. However, when there are enormous number of feature points, the calculation work of the brute force matching approach becomes quite huge, making it difficult to fulfil the real-time needs of SLAM. To solve to problem of calculation burden, the FLANN (Fast Library for Approximate Nearest Neighbors) algorithm [28] is applied to situations with enormous feature points. The FLANN algorithm creates an index structure for the descriptors of feature points. For the feature points to be matched, the FLANN algorithm will perform an approximate search in the index structure. Determine whether there is a match by comparing the similarity (such as Hamming distance) between the point to be matched and the nearest neighbor point. FLANN is an approximate matching algorithm, so the matching accuracy is not as high as violent matching. However, when there are too many feature points, the fast and effective matching provided by the FLANN algorithm is far superior to the negative effect its inaccuracy has.

#### 2.1.1.3　　Motion estimation

The matched feature point pairs obtained after feature extraction and matching can be used for camera motion estimation. Depending on the type of camera (monocular, binocular, RGD-D), the motion estimation problem can be solved in different ways:

- Monocular camera: Only the two-dimensional coordinates of key points can be obtained, so the motion is estimated through two sets of two-dimensional points, which can be solved using epipolar geometry.
- Binocular, RGB-D camera: can obtain the two-dimensional coordinates and distance information of key points, thus estimating motion through three-dimensional points. This problem is solved with ICP.

### 2.1.2 Optimization

The motion trajectory and pose information calculated by VO gradually increase over time. Due to the objective error of the sensor and the calculation error of VO, the cumulative error continues to increase during the long-term movement of the robot. Therefore, the backend Optimization part needs to continuously optimize the frontend to reduce cumulative errors.

#### 2.1.2.1 Key frame selection

In the continuous movement of the robot, there exists great repeatability between two adjacent frames, and optimizing each frame will bring an extremely large and useless computational burden to computer. Therefore, keyframes need to be selected from consecutive frames for optimization. The selection of keyframes needs to meet the following two requirements:

- There are extensive feature points on key frames, which can well represent the surrounding environment information.
- There needs to be a certain correlation between key frames. There must be extensive different feature points to avoid system redundancy, and some of the same feature points need to be retained to ensure mutual constraints.

The keyframe selection method of ORB-SLAM [5] is taken as an example here. ORB-SLAM selects new keyframes when the following requirements are met:

- More than 20 frames have passed since the last global relocation.
- The local map construction is currently idle, or there have been more than 20 frames since the last keyframe was inserted.
- The current frame tracks less than 50 map cloud points.
- The current frame tracks less than 90% of the reference keyframe cloud points.

#### 2.1.2.2 Pose estimation

There are two basic ways to create a global optimization problem to conduct pose estimation：

- **KF (Kalman Filter) based on Markov property**: Markov property means that "the current state only depends on the state at the previous moment". This property is also used in the reinforcement learning problem introduced in the next section. The KF for Visual SLAM problems (nonlinear) gives the greatest posterior estimate using a single linear approximation, or the result of one iteration in the optimization process.

- **Nonlinear optimization**: Nonlinear optimization optimizes the sensor poses and the spatial position of each feature point by considering the current and previous states.

Nonlinear optimization has a wider scope than KF and is more beneficial to global optimization. The early Visual SLAM system, such as MonoSLAM, uses KF as the backend. After researchers discovered the many disadvantages of using KF as the backend, later Visual SLAM including PTAM and ORB-SLAM gradually takes nonlinear optimization as the replacement. BA (Bundle Adjustment) optimization [29] is the method commonly used to optimize the poses of key frames. BA optimization optimizes the reprojection error while completing the optimization of the camera pose by reprojecting the beam composed of three-dimensional space points and feature points on the imaging plane.

### 2.1.3 Loop Closure Detection

Loop Closure Detection introduces closed-loop constraints by detecting the coincidence of the same or similar scenes at different time points during the camera's data collection process to reduce cumulative errors and improve the accuracy of mapping and positioning.

Loop Closure Detection methods can be divided into odometry-based method and appearance-based method. Odometry-based method considers detecting whether there is a loop closure relationship when the camera returns to the vicinity of a previous position. However, due to the objective existence of cumulative errors, this method is not valid. Appearance-based method detects loop closure based only on image similarity. This approach eliminates the effects of cumulative errors and is applicable to a variety of environmental conditions. At present, appearance-based method is the mainstream method for Loop Closure Detection.

How to compare the similarity between images? The direct way is to represent the two images as matrices, and then subtract the two matrices. The result matrix can be used to measure the similarity between two images. This method forgets to consider the influence of light condition and camera angle on images. The pixel grayscale of an image is an unstable value and will be affected by lighting. Different lighting conditions in the same scene will lead to huge difference in the grayscale of the image. In addition, changes in the camera angle will also cause huge changes in the grayscale of the image. Therefore, this method is not practical, and BoW (Bag of Word) is the solution.

In Visual SLAM, the types of features and the number of their occurrences in an image are used to describe the image instead of its pixel grayscale. The image features here are represented by single words in BW. The main work of the Loop Closure Detection based on Bag of Word is divided into three steps:

- Cluster the external environment scenes to generate a general dictionary.
- Extract image features for each input keyframe and mapping them through the dictionary to obtain its Feature vector.
- Perform loop closure detection based on the feature vector of each keyframe.

The BoW generation problem is a type of clustering problem and can be implemented using the classic K-means algorithm of unsupervised learning [30].

With the rise of CNN (Convolutional Neural Network), loop closure detection algorithms based on deep learning have become the research hotspot in Visual SLAM [31]. Compared with BoW-based algorithms, DL-based algorithms can automatically learn image features through neural network models without the need for manually designed feature extraction algorithms and can learn richer contextual information.

### 2.1.4 Visual SLAM system used for path planning

#### 2.1.4.1 ORB-SLAM3 with dense map construction

The Visual SLAM system of this project is developed based on the open-source ORB-SLAM3 project [7]. ORB-SLAM3 is the pinnacle of current feature-point-extraction-based Visual SLAM. Its code is clear and easy to read, and has complete comments, which facilitates the development of projects based on it. However, ORB-SLAM also has drawbacks. The mapping of ORB-SLAM uses sparse feature points, which can only afford the positioning function. For more complex requirements including navigation, obstacle avoidance, etc., the extremely limited environmental information provided by sparse feature points cannot be satisfied. Therefore, this project expands the functions of ORB-SLAM3 to meet the requirement of path planning. Taking advantage of the depth image information obtained by the RGB-D camera, this project adds a dense point cloud mapping thread to ORB-SLAM3. The resulting dense point cloud map contains very rich information, which far exceeds the needs of path planning. Redundant information will increase the storage space of the point cloud map, and what is worse is that it will greatly increase the ineffective calculation cost. Therefore, this project uses a more lightweight octomap to replace the dense point cloud map.

#### 2.1.4.1 Map conversion

The octree is a hierarchical data structure, with the entire three-dimensional space as the root node, continuously divided, and each parent node has eight child nodes. Through recursive division, an octree structure with different levels can be constructed. The smallest node size is the resolution size, as shown in figure 3:

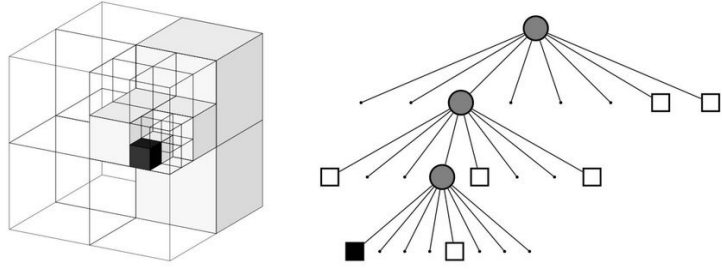

**Figure 3.** Example of an octree map with free (white) and occupied (black) cells [32].

When converting a dense point cloud map to octomap, the basic idea is to traverse the point cloud data and assign each point to the corresponding octree node. This is accomplished by comparing the 3D coordinates of the point with the region of space represented by the octree node. For each octree node, the division can be stopped if the number of point clouds it contains exceeds a preset threshold or reaches the maximum depth. Otherwise, continue to recursively subdivide the node into eight sub-nodes and assign the corresponding point cloud to the sub-nodes.

Since the scenario this project considers is robot moving on a plane, the three-dimensional information of Octomap is not very important. Through a simple projection transformation, the Octomap is converted into a two-dimensional grid-based map for final path planning. Such a series of map conversions shown in figure 4 reduces the complexity of the map information and improves the efficiency of algorithm implementation.

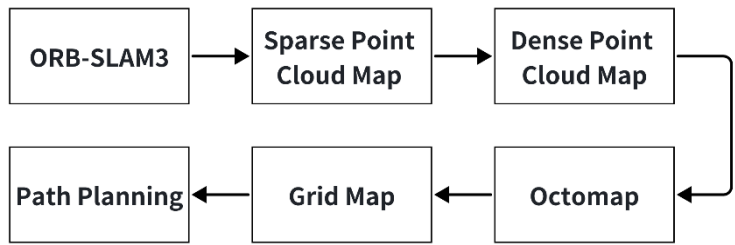

**Figure 4.** Map conversion process in this project.

## 2.2 Research on path planning based on RL

Path planning algorithm is one of the key technologies for mobile robots to complete autonomous navigation. After obtaining the map of the surrounding environment through front-end Visual SLAM, the path planning algorithm can perform path planning for the robot to complete autonomous navigation based on the obtained map information. As what has been mentioned in Introduction part, the research focus of this project is path planning algorithm based on RL. Therefore, this section will introduce the three RL algorithms (Q-learning, SARSA, DQN) used in the experiments and explain how to use them for path planning and will not go into details about the traditional path planning algorithm.

## 2.2.1 Reinforcement learning principles

The core principle behind reinforcement learning is to make agent learn from the interaction with its surrounding environment. The agent adjusts its policy according to rewards in the process of continuous trial and error to maximize cumulative future reward. The purpose of RL is to identify the best policy that allows the agent to get the highest cumulative reward from its interactions with the environment. The RL model mainly includes five elements: agent, environment, state, action and reward, as shown figure 5:

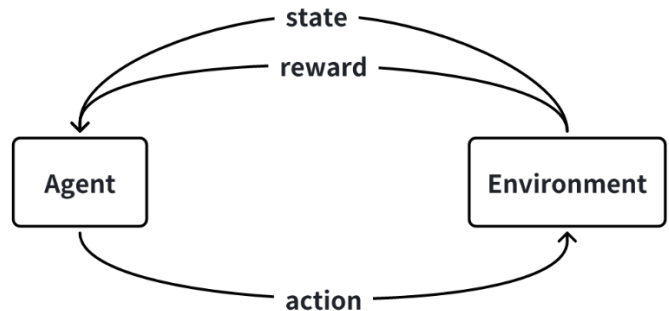

**Figure 5.** Reinforcement learning model.

- Agent: The central part of RL and becomes intelligent through continuous interaction with the environment.
- Environment: The abstract collection of the external environment where the agent is located.
- State: Summary of the current environment and the agent needs to choose actions according to the current state.
- Action: The decision made by the agent according to the current state and the agent interact with the external environment through actions. Different states leads to different actions and actions can affect future state.
- Reward: A real number used to access the quality of the action conducted by the agent in the current state. Positive rewards are given to the good actions and negative rewards are given for bad actions.

MDP (Markov Decision Process) is a commonly used modelling tool for RL. A MDP can be described using a five-tuple $(S, A, P, R, \gamma)$:

- $S$: State space, the set of all possible states.
- $A$: Action space, the set of all possible actions.

- $P$: State transition function indicates the probability that the agent transfer to the next state $s_{t+1}$ after performing action $a_t$ in current state $s_t$, which is expressed as $P(s_{t+1}|s_t, a_t)$.
- $R$: Reward function, the instant reward gained by the agent after performing action $a_t$ in current state $s_t$.
- $\gamma$: Discount rate, a real number between 0 and 1 used to balance the importance of current reward and future reward.

State transition satisfies the Markov property is a prerequisite for MDP, that is, the next state only depends on current state and current action and is unrelated to past states and actions. This property ensures that the agent can predict the future state through the current state and action.

The goal of RL is to obtain an optimal policy that maximizes cumulative future reward, also called return, rather than maximizing the current reward. The cumulative future reward is not simply the addition of current reward and future rewards. The future reward is considered as less important than current reward. Therefore, the concept of discount rate is introduced in the cumulative future reward. The discounted return $U_t$ can be defined as:

$$U_t = R_t + \gamma \cdot R_{t+1} + \gamma^2 \cdot R_{t+2} + \gamma^3 \cdot R_{t+3} + \cdots \tag{1}$$

It is obvious that the discounted return $U_t$ can be used to measure the quality of the current state. The greater the return, the better the current state. However, the future rewards in equation (1) are unknown in current state. The expectation of $U_t$ can eliminate the uncertainty, so it is used to measure the quality of current state instead. This expectation is called the value function. Value functions can be divided into different types according to its functions. The following is a brief introduction to three kinds of value functions: action-value function $Q_\pi(s, a)$, optimal action-value function $Q_*(s, a)$, and state-value function $V_\pi(s)$. To avoid confusion, in the remainder of this section uppercase letters $(S, A)$ represent variables and lowercase letters $(s, a)$ represent actual observations.

**Action-value function $Q_\pi(s, a)$**

The uncertainty of $U_t$ comes from all the future state-action pairs:

$$S_{t+1}, A_{t+1}, S_{t+2}, A_{t+2}, \cdots, S_n, A_n \tag{2}$$

The action-value function can be obtained by calculating expectation for all future action-state variables:

$$Q_\pi(s_t, a_t) = \mathbb{E}_{S_{t+1}, A_{t+1}, \cdots, S_n, A_n}[U_t | S_t = s_t, A_t = a_t] \tag{3}$$

$Q_\pi(s_t, a_t)$ depends on current state $s_t$, current action $a_t$ and policy function $\pi$. The policy here can be deterministic policy or random policy. For random policy, the policy function is a probability density function:

$$\pi(a|s) = \mathbb{P}(A = a|S = s) \tag{4}$$

This indicates the probability for agent to choose action $a$ in state $s$. For deterministic policy, the action $a$ given by policy is determined:

$$\pi(s) = a \tag{5}$$

Action-value function can be used to access the quality of current state, actions and strategies.

**Optimal action-value function $Q_*(s, a)$**

Based on action-value function, optimal action-value function excludes the influence of strategy:

$$Q_*(s_t, a_t) = \max_\pi Q_\pi(s_t, a_t) \tag{6}$$

Only consider the optimal policy function $\pi^*$:

$$\pi^* = \mathop{argmax}_\pi Q_\pi(s_t, a_t) \tag{7}$$

$Q_*(s_t, a_t)$ depends on current state $s_t$ and current action $a_t$ and is unrelated to policy.

**State-value function $V_\pi(s)$**

What if agent only wants to the quality of current state without considering the impact of actions? State-value function $V_\pi(s)$ consider the current action as a variable $A_t$, calculates expectations of it, and eliminates its impact on the value function:

$$V_\pi(s) = \mathbb{E}_{A_t \sim \pi(\cdot|s_t)}[Q_\pi(s_t, a_t)] = \sum_{a \in \mathcal{A}} \pi(a|s_t) \cdot Q_\pi(s_t, a) \tag{8}$$

State-value function only depends on policy $\pi$ and current state $s_t$ and is unrelated to current action.

RL algorithms can be divided into value-based learning and policy-based learning according to the object of learning. Value-based learning selects actions according to the value function while the actions in policy-based learning is selected according to the policy function. The three algorithms introduced in this section all belong to value-based learning.

### 2.2.1 Q-learning

Given the current state $s_t$, the optimal action-value function $Q_*$ can score the action, for example, when the action space size is 3:

$$Q_*(s_t, a_1) = -100 \quad Q_*(s_t, a_2) = 0 \quad Q_*(s_t, a_3) = 100 \quad a_1, a_2, a_3 \in \mathcal{A} \quad |\mathcal{A}| = 3$$

It is obvious that $a_3$ is the optimal action under current state $s_t$. The objective of Q-learning is to learn the optimal action-value function $Q_*$ and choose action according to equation (9). Q-learning algorithm and the SARSA algorithm in the next subsection are both TD ( Temporal Difference) algorithms, which enable the agent to obtain the maximum cumulative reward in the environment by gradually updating the value function.

$$a_t = \underset{a \in \mathcal{A}}{\operatorname{argmax}} Q_*(s_t, a) \tag{9}$$

When the action space $\mathcal{A}$ and state space $\mathcal{S}$ are both finite sets, Q-learning can use a Q-table to approximate $Q_*$. Table 2 shows an example Q-table for discrete action space ($a_1$, $a_2$, $a_3$) and discrete state space ($s_1$, $s_2$, $s_3$). All the elements in the table are initialized as 0 first and each time one element of the table is updated, the table will eventually converge to $Q_*$.

**Table 2.** Simple Q-table example

| State \ Action | $a_1$ | $a_2$ | $a_3$ |
|---|---|---|---|
| $s_1$ | $Q_*(s_1, a_1)$ | $Q_*(s_1, a_2)$ | $Q_*(s_1, a_3)$ |
| $s_2$ | $Q_*(s_1, a_1)$ | $Q_*(s_2, a_2)$ | $Q_*(s_2, a_3)$ |
| $s_3$ | $Q_*(s_1, a_1)$ | $Q_*(s_3, a_2)$ | $Q_*(s_3, a_3)$ |

Considering the definition of the return $U_t$, equation (1) can be rewrote as:

$$U_t = R_t + \gamma \cdot \sum_{k=t+1}^{n} \gamma^{k-t-1} \cdot R_k \tag{10}$$

$\sum_{k=t+1}^{n} \gamma^{k-t-1} \cdot R_k$ is equal to $U_{t+1}$; therefore, equation (10) can be expressed as:

$$U_t = R_t + \gamma \cdot U_{t+1} \tag{11}$$

According to equation (3), equation (6) and equation (11), the optimal Bellman equation can be expressed as:

$$Q_*(s_t, a_t) = \mathbb{E}_{S_{t+1} \sim p(\cdot|s_t, a_t)}[R_t + \gamma \cdot \underset{A \in \mathcal{A}}{\max} Q_*(S_{t+1}, A) \,|\, S_t = s_t, A_t = a_t] \tag{12}$$

When agent performs action $a_t$ under current state $s_t$, it can observe current reward $r_t$ and next state $s_{t+1}$, which can be described with a four-tuple $(s_t, a_t, r_t, s_{t+1})$. Therefore, equation (12) can be expressed as:

$$r_t + \gamma \cdot \max_{a \in \mathcal{A}} Q_*(s_{t+1}, a) \tag{13}$$

Equation (13) is the Monte Carlo approximation [33] of equation (12); therefore, TD target can be expressed as:

$$TD_{target} = r_t + \gamma \cdot \max_{a \in \mathcal{A}} Q_*(s_{t+1}, a; w) \tag{14}$$

Compared with $Q_*(s_t, a_t)$, $TD_{target}$ is partially based on truly observed rewards. Therefore, $TD_{target}$ is considered as more reliable. The element at position $(s_t, a_t)$ should be updated to approach $TD_{target}$:

$$Q_{table}(s_t, a_t) = (1 - \alpha) \cdot Q_{table}(s_t, a_t) + \alpha \cdot TD_{target} \tag{15}$$

$\alpha$ is the learning rate.

### 2.2.2 SARSA

SARSA algorithm has many similarities to Q-learning algorithm and the key difference is that the target of SARSA is learning action-value function $Q_\pi(s, a)$. SARSA is the abbreviation of state-action-reward-state-action. This is because it uses the five-tuple $(s_t, a_t, r_t, s_{t+1}, a_{t+1})$, in which $a_{t+1}$ is obtained according to policy function $\pi(\cdot | s_{t+1})$. Therefore, the optimal Bellman equation used in Q-learning is replaced with Bellman equation for SARSA:

$$Q_*(s_t, a_t) = \mathbb{E}_{S_{t+1} \sim p(\cdot | s_t, a_t)}[R_t + \gamma \cdot Q_\pi(S_{t+1}, A) | S_t = s_t, A_t = a_t] \tag{16}$$

The subsequent update process is consistent with Q-learning.

**Comparison between Q-learning and SARSA**

Before the comparison of Q-learning and SARSA, concepts about on-policy and off-policy need to be explained. In the learning process of RL, the agent interacts with the environment through the behaviour policy to collect experience, which control the agent in the interacting process. The target of RL is to learn the policy function, which is known as the target policy. For on-policy algorithm, the behaviour policy and the target policy are the same while for off-policy algorithm, they are different.

Q-learning aims to learn $Q_*$ and Q-learning does not depend on policy. Whatever the behaviour policy of Q-learning is, it will not affect the training of its target policy. SARSA aims to learn $Q_\pi$ and different policy will lead to different $Q_\pi$. Therefore, the behaviour policy and target policy of SARSA must be the same. Based on what has been introduced, Q-learning is off-policy and SARSA is on-policy.

On-policy and off-policy has different training techniques. For off-policy algorithms, experience replay [34] can be used to make the experience samples independent to each other and reuse the collected experience. However, experience replay is not appliable to on-policy algorithm like SARSA. In the experiment section, experience replay is used in Q-learning and SARSA.

### 2.2.3 DQN

In actual applications, the action space and state space are complex, which makes it difficult of approximate $Q_*$. Therefore, deep Q-network $Q(s, a; w)$ is often used to approximate $Q_*(s, a)$. The $w$ represents the parameters of the neural network. Through continuous update, $Q(s, a; w)$ will be as close as possible to $Q_*(s, a)$. The input of $Q(s, a; w)$ is the current state, and the output of $Q(s, a; w)$ is an $|\mathcal{A}|$-dimensional vector, containing the score of actions used in the current state. The higher the score, the better the action.

Like Q-learning, DQN also uses TD algorithm for training. The loss function of the training process is calculated based on TD error. TD error is the difference between current prediction and TD target:

$$TD_{error} = Q_*(s_t, a_t; w) - TD_{target} \qquad (17)$$

Commonly used loss function is MSE (Mean Square Error):

$$L(w) = \frac{1}{2} TD_{error}^2 \qquad (18)$$

The gradient of $L(w)$ with respect to $w$:

$$\nabla_w L(w) = TD_{error} \cdot \nabla_w Q_*(s_t, a_t; w) \qquad (19)$$

Use gradient descent to update $w$ to make the current prediction closer to TD target:

$$w = w - \eta \cdot TD_{error} \cdot \nabla_w Q_*(s_t, a_t; w) \qquad (20)$$

$\eta$ is the learning rate.

### 2.2.4 Path planning based on RL

Path planning based on RL requires system modelling of the path planning problem using RL. As introduced in section 2.2.1, the five basic elements in RL need to be determined first: Agent, Environment, Action, Reward, and State. Then, appropriate RL algorithms should be applied to solve the path planning problem. In the path planning scenario, the robot is the Agent and the surrounding environment where the robot is located is the Environment. The robot interacts with the environment each time by selecting an action according to the output of the RL algorithm used. According to the state of the robot after performing the action, the environment returns a reward to the robot. In the process of interacting with the environment, the robot continuously updates the reinforcement learning algorithm to obtain the maximum cumulative reward. This project has used the three value-based algorithms introduced before: Q-learning, DQN, SARSA and has conducted comparative experiments among them. The detailed experimental design will be introduced in Section 3.

# 3 Experimental design and results

## 3.1 Visual SLAM simulation

This part of experiment is based on the improved ORB-SLAM3 of this project. This part of the experiment is divided into two parts according to the source of the dataset:

- The first part uses the open-source TUM Dataset to verify the feasibility of the ORB-SLAM3 system with added dense point cloud mapping function.

- In the second part, the self-made dataset is used to verify the applicability of the system. The part includes the experimental environment setup and dataset recording using RGB-D camera.

All the simulations in this part were conducted on Ubuntu 18.04 system with ROS (Robot Operation System) noetic installed.

### 3.1.1 Simulation based on TUM Dataset

#### 3.1.1.1 TUM Dataset

The TUM Dataset [35] is a series of open-source data sets used in the fields of computer vision and robotics maintained by Technical University of Munich (TUM). These datasets contain a variety of different scenes and objects and are used to train and test various vision and robotics algorithms. The experiment in this section uses the RGB-D SLAM Dataset in the TUM data set for the dense point cloud mapping experiment of ORB-SLAM3. The resulting dense point cloud map will be converted into the octomap and grid map in real time. The dataset is composed of the color and depth images of a Microsoft Kinect sensor along its ground-truth trajectory. The sequence chosen by this project was record form the RGB-D camera mounted on top of a Pioneer robot moving around a larger maze full of various obstacles (tables, walls, chairs, etc.). This complex experimental environment is closer to the real environment where robots are used in real life.

#### 3.1.1.2 Experimental results

Figure 6 shows the experimental results based on the RGB-D SLAM Dataset. Figure 6.a shows the dense point cloud map constructed from the improved ORB-SLAM3 system proposed by this project. The dense cloud map contains abundant information of the experimental environment. The pink trajectory in figure 6.a is the ground-truth trajectory of the robot, which indicates the robot has finished a loop in the environment. Figure 6.b and figure 6.c show the converted octomap and grid map of the dense point cloud map. The white areas on the grid map represent positions that are not occupied by obstacles and can be passed by the robot. Black areas represent areas occupied by obstacles that the robot cannot pass. After the map conversion, the size of the grid map is only 4.34KB. The grid map contains enough information required for path planning.

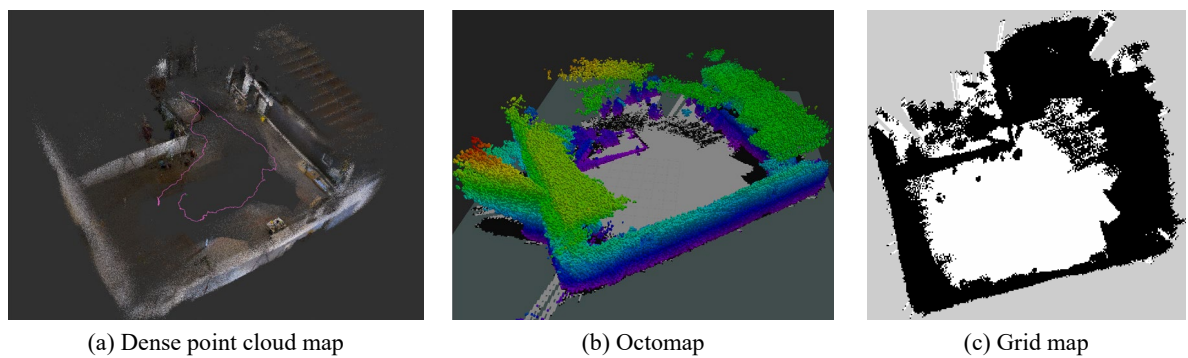

(a) Dense point cloud map        (b) Octomap        (c) Grid map

**Figure 6** Results for dense point cloud mapping and map conversion

### 3.1.2 Simulation based on self-made dataset

#### 3.1.2.1 Experimental Environment

The experimental environment is a simple maze with a scale of 240*240 cm shown in figure 7. This maze is made of 40*30*10 cm cardboard boxes. The entrance in the lower left corner is the starting point of the robot, and the exit in the upper right corner is its target point. Each cardboard box is labeled with different colored A4 papers. The main purpose of this is to add corner points to facilitate the feature extraction by the ORB-SLAM3 system. In the initial experiments, when there was no colored paper on the cardboard boxes, the effect of feature extraction was poor due to the small difference in grayscale in the image, resulting in failure of localization and mapping. Therefore, the subsequent dataset recording experiments are all performed in the environment shown in figure 7.

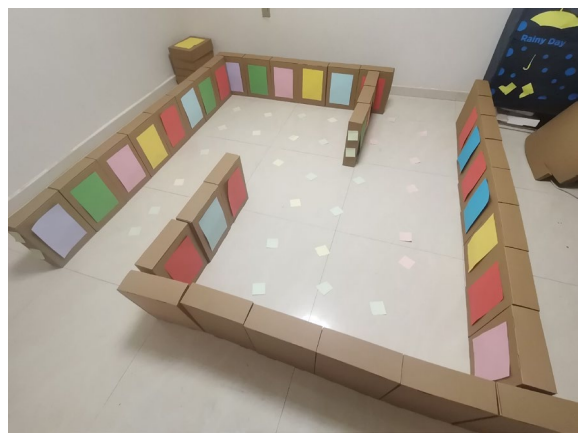

**Figure 7.** Experimental environment for dataset recording

### 3.1.2.2 Dataset recording

The RGB-D camera used for dataset recording is Astra Pro Plus from Orbbec, as shown in figure 8. This RGB image of the camera supports up to 1920*1080@30fps, and the depth image of it supports up to 640*480@30fps. During the experiment, the resolution and frame rate set for color images and depth images were both 640*480@30fps. The Astra Pro Plus camera captures RGB image through a monocular camera and obtains depth image through the reflection time from the projected infrared pulse.

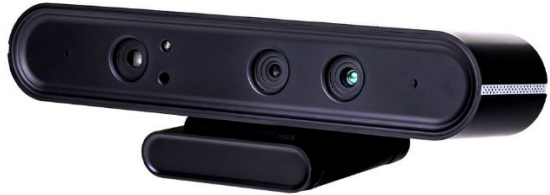

**Figure 8.** Astra Pro Plus RGB-D camera

Camera Calibration of Astra Pro Plus can be conducted using the ROS packages provided by Orbbec. Camera Calibration can obtain the Camera Intrinsics and Distortion Coefficients, which are important for calculating the corresponding relationship between each pixel of the camera and the real-world location. The Intrinsic Matrix and Distortion Coefficients of the camera used for dataset recording are shown in equation (21) and (22). These parameters are used in the experiment based on the self-made dataset to increase the accuracy of localization and mapping of ORB-SLAM3.

$$\begin{bmatrix} 533.712968 & 0 & 318.091435 \\ 0 & 534.001252 & 241.847649 \\ 0 & 0 & 1 \end{bmatrix} \quad (21)$$

$$\begin{aligned} k_1 &= 0.231222 \\ k_2 &= -0.784899 \\ p_1 &= -0.003257 \\ p_2 &= -0.000105 \\ k_3 &= 0.917205 \end{aligned} \quad (22)$$

During the experiment, the camera was held 20 cm above the ground to simulate that the camera is mounted on the robot. Keep this height and move from the starting point of the maze to the starting point to complete the recording of the dataset.

### 3.1.2.3 Experimental results

The experimental results are shown in figure 9 and 10. Figure 9.a and figure 9.b show the RGB image and the corresponding Depth image captured by the camera in the dataset recording process. Figure 9.c shows the feature point extraction results by ORB-SLAM3, in which the green points represent the feature points extracted from the image. Most of the feature points located at the edges of the colored A4 papers. The result shows that adding the colored paper helps to increase the feature point.

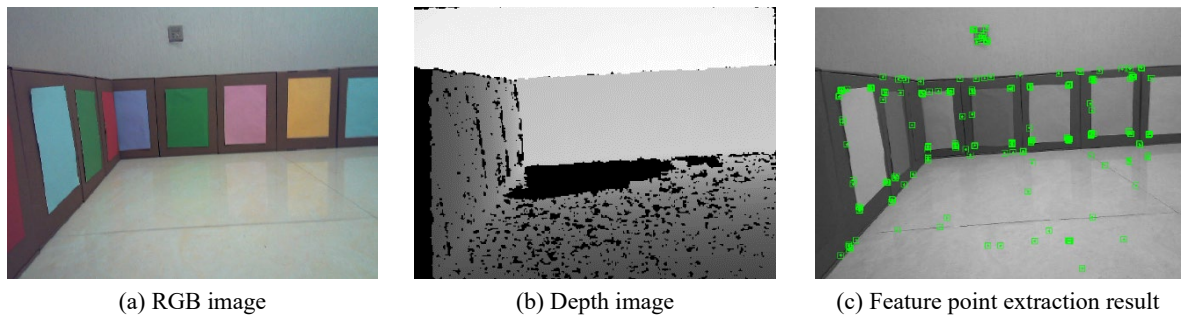

(a) RGB image　　　　　　　(b) Depth image　　　　　　(c) Feature point extraction result

**Figure 9.** The images captured by the camera and the feature extraction result

Figure 10 shows the dense point cloud mapping result and the map conversion results. Figure 10.a shows the dense point cloud map of the experimental environment, which contains abundant information of the environment shown in figure 7. The pink trajectory shows the ground-truth trajectory of the camera form the starting point to the target point. Figure 10.b and figure 10.c shows the converted octomap and grid map of the dense cloud map. The grid map used for path planning only takes up 3.75KB in size. The map conversion operation removes a large amount of redundant information and retains the necessary information required for path planning.

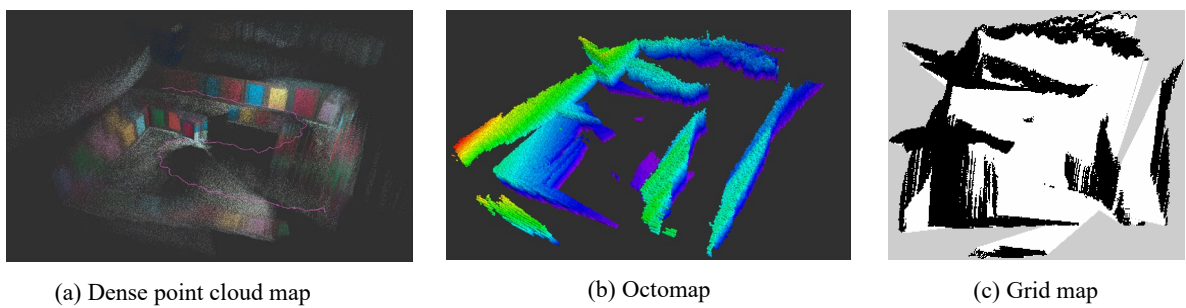

(a) Dense point cloud map　　　　　(b) Octomap　　　　　　(c) Grid map

**Figure 10.** Results for dense point cloud mapping and map conversion

## 3.2 Path planning simulation

### 3.2.1 Simulation environment

For the path planning experiment, the environment uses the grid map mentioned in Section 2.1.4 to simulate the two-dimensional external environment generated by the front-end Visual SLAM part. Figure 11 is the grid map model generated during the experiment, in which the two blue rectangles represent obstacles, and the two small red squares represent the starting point and target point of the robot respectively (the starting point is in the lower left corner and the target point is in the upper right corner). In the simulation experiment, the projected shape of the robot on the two-dimensional plane was viewed as a $10 * 10$ square, the size of the entire grid map was $100 * 100$, and the size of the two obstacles was $10 * 60$.

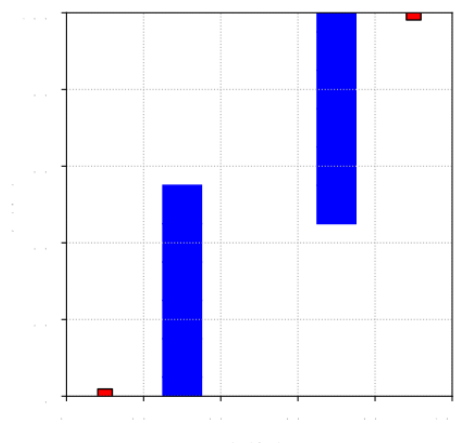

**Figure 11.** Simulation environment for path planning

### 3.2.2 Experiment design

This section shows the design of the basic elements in MDP model of RL based on the simulation environment shown in the last section. The main design content includes the following parts:

**State space**

In the grid map environment, the surrounding environment of the agent mainly includes three elements: map boundaries, obstacles, and the target point. These elements, including the agent, have two-dimensional coordinates in the grid map. Based on their coordinate and scale information, it can be determined whether the agent has reached the target point or collided with the boundary or obstacle. Therefore, the state of the agent can be represented by its two-dimensional coordinates, and the state space is defined as follows:

$$S = [x, y] \tag{23}$$

**Action space**

As introduced in the Section 2.2.1, action space is the set of all possible actions of the agent. In order to reduce the complexity of the experiment and facilitate algorithm training, the experimental design of this project did not consider the complex motion of the robot, including continuous movement angles and continuous movement distances. This project only considers the robot moving with fixed distances in discrete directions. As shown in equation (24) and figure 12, the action space of the agent is defined as eight discrete actions: up ($a_1$), down ($a_2$), left ($a_3$), right ($a_4$), upper left ($a_5$), upper right ($a_6$), lower left ($a_7$), lower right ($a_8$). Each action leads to the change of the agent's state as shown in equation (25), in which $d$ is set as 10 in the experiment.

$$\mathcal{A} = [a_1, a_2, a_3, a_4, a_5, a_6, a_7, a_8] \tag{24}$$

$$S_{t+1} \leftarrow \begin{cases} S_t + [0, d] & a_t \epsilon a_1 \\ S_t + [0, -d] & a_t \epsilon a_2 \\ S_t + [-d, 0] & a_t \epsilon a_3 \\ S_t + [d, 0] & a_t \epsilon a_4 \\ S_t + [-d, d] & a_t \epsilon a_5 \\ S_t + [d, d] & a_t \epsilon a_6 \\ S_t + [-d, -d] & a_t \epsilon a_7 \\ S_t + [d, -d] & a_t \epsilon a_8 \end{cases} \tag{25}$$

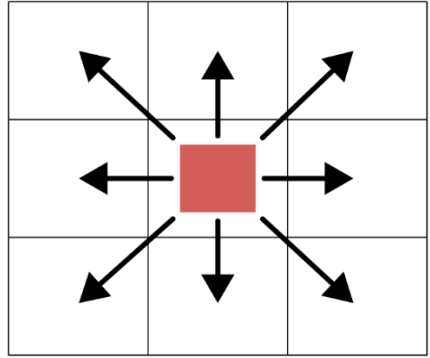

**Figure 12.** Possible actions of the agent

**Reward function**

The reward function can be used to measure the result of the agent performing the action $a_t$ in the current state $s_t$. In the grid map environment, the agent has four kinds of states: collision with obstacles or the map boundary $S_1$, the action chosen belongs to $[a_1, a_2, a_3, a_4]$ and no collision $S_2$, the action chosen belongs to $[a_5, a_6, a_7, a_8]$ and no collision $S_3$, and reaching the target point $S_4$. $S_1$ is the state that the agent should avoid during movement, so when the agent is in this state, it should be given a negative reward. The goal of path planning is to find the shortest path from the starting point to the target point. Therefore, if there is no collision, the number of $S_2$ and $S_3$ occurrences should be as small as possible. The agent would receive a negative reward when its state belongs to $S_2$ or $S_3$. Besides, $S_3$ gives a larger punishment because the agent has moved for longer distance. $S_4$ is the target state of the agent. Therefore, when the agent reaches $S_4$, it will receive a positive reward. To sum up, the reward function can be represented by a piecewise function:

$$R = \begin{cases} -20 & S \in S_1 \\ -1 & S \in S_2 \\ -1.5 & S \in S_3 \\ 20 & S \in S_4 \end{cases} \quad (26)$$

When the agent collides with the map boundary or obstacles, it gets a reward of -20; when the agent moves horizontally or vertically and does not collide, it gets a reward of -1; when the agent moves in a diagonal direction and does not collide, it gets a reward of -1.5; when the agent reaches the target point, it gets a reward of 20.

**Policy**

The agent selects actions based on its policy in different states. Q-learning, DQN, and SARSA all score actions and agent select the optimal action according to the scores in the current state. However, such the deterministic policy lacks exploratory. If it only chooses the optimal action in each state, the agent may fall into the local optimal solution and be unable to discover new, potentially optimal policies. This project chooses the ε-greedy as the exploration policy. It helps keep a balance between exploration and exploitation during training. When using the ε-greedy policy for action selection, actions are randomly selected with ε probability, and the optimal action is selected with 1-ε probability. This can effectively increase the exploratory and avoid falling into the local optimal solution. However, the lack of clear directionality of the ε-greedy strategy also makes it unable to guarantee global optimality. In addition, too much exploration by the agent will cause the algorithm to converge too slowly, wasting training time and resources. Therefore, when applying the ε-greedy strategy in this project, the exploration rate ε gradually becomes smaller as the training episodes increase, as shown in equation (27).

$$\boldsymbol{\varepsilon} \leftarrow \boldsymbol{\varepsilon} - \boldsymbol{\Delta} \quad (27)$$

The effect of this processing is: in the early stage of training, when ε is large, the agent has larger probability to choose action randomly to do exploration. As the training progresses, ε gradually decreases, and the agent will have larger possibility to choose optimal actions. This not only increases exploratory, but also ensures global optimality.

**Parameter configuration**

The parameters used in the experiment is set as table 3:

Table 3. Parameter configurations in the experiment

| Parameter | Value |
| --- | --- |
| Discount factor, $\gamma$ | 0.99 |
| Initial exploration rate, $\varepsilon_{initial}$ | 0.6 |
| Final exploration rate, $\varepsilon_{final}$ | 0.1 |
| Replay buffer size | 100000 |
| Batch size | 128 |
| Total Number of episodes | 300 and 500 |
| Learning rate of Q-learning and SARSA, $\alpha$ | 0.01 |
| Learning rate of DQN, $\eta$ | 0.001 |

With the increase of the number of the training episode, the initial exploration rate will gradually decrease to the final exploration rate.

Experience replay is used for the training of both Q-learning and DQN, which are off-policy algorithms. The replay buffer size is set as 100000 and batch size is 128. 128 samples are selected randomly each time form the replay buffer to update the algorithms.

The total number of episodes required for convergence was determined through empirical experience. For the DQN algorithm, a total of 300 episodes were found to be sufficient for convergence, while for Q-learning and SARSA, 500 episodes were required. To observe the convergence behavior, testing experiments were conducted by gradually decreasing the total number of episodes by 100, starting from an initial value of 1000 episodes in each experiment. The convergence of the algorithms was assessed based on their performance during these experiments.

Through the analysis of the experimental results, it was observed that DQN achieved convergence when the total number of episodes was reduced to 300, while Q-learning and SARSA converged at 500 episodes. The choice of 300 episodes for training DQN and 500 episodes for training Q-learning and SARSA was made based on these convergence points. These values were selected to strike a balance between achieving convergence within a reasonable number of episodes and ensuring optimal learning performance for each algorithm.

Learning rate of DQN and Q-learning mentioned in Section 2.2.1 and the learning rate of DQN mentioned in Section 2.2.3 are different parameters, which should not be confused.

### 3.2.3 Experimental results

The experimental results of Q-learning, DQN, and SARSA are shown in figure 13, figure 14, and figure 15 respectively. The experimental results include accumulated reward per episode, number of steps taken by the agent per episode, and the optimal path obtained by each algorithm after the training process. These results show the convergence of the algorithm during the training process and the final performance.

From the line graphs in figure 13.a and figure 13.b, the accumulated reward and number of steps taken per episode of Q-learning both converge to stable values after 500 training episodes. As the optimal path shown in figure 13.c, Q-learning successfully find a path from the starting point to the target point without collision. The number of steps and the accumulated reward of the optimal path are 18 and -2 respectively.

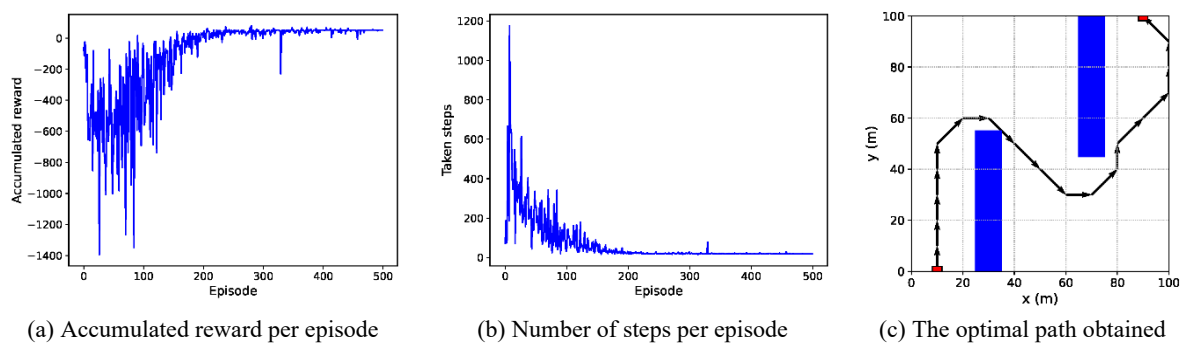

(a) Accumulated reward per episode    (b) Number of steps per episode    (c) The optimal path obtained

**Figure 13.** Experimental results of Q-learning

From the line graphs in figure 14.a and figure 14.b, the accumulated reward and number of steps taken per episode of DQN both converge to stable values after 300 training episodes. As the optimal path shown in figure 14.c, Q-learning successfully find a path from the starting point to the target point without collision. The number of steps and the accumulated reward of the optimal path are 18 and -1 respectively.

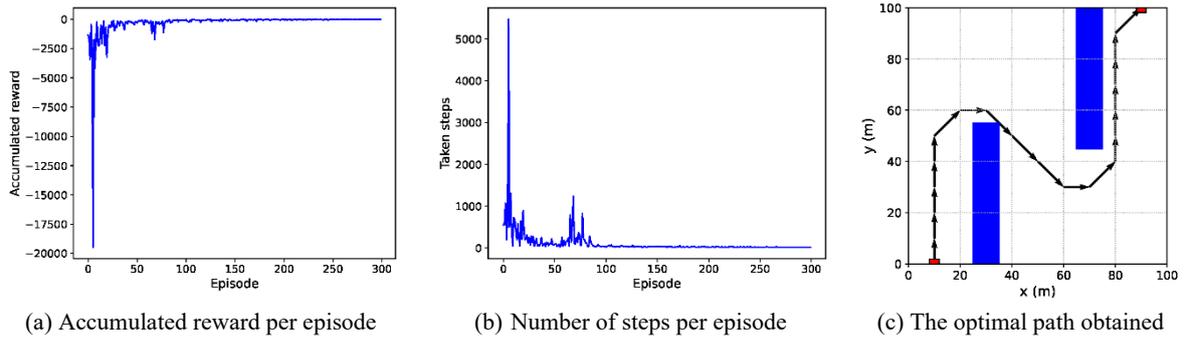

(a) Accumulated reward per episode  (b) Number of steps per episode  (c) The optimal path obtained

**Figure 14.** Experimental results of DQN

Line graphs in figure 15.a and figure 15.b exhibit pronounced fluctuations. Both the accumulated reward and number of steps taken per episode fail to converge to stable values after 500 training episodes, indicating the significant oscillations in the training process of SARSA algorithm. Contrasted with figure 13.c and figure 14.c, figure 15.c depicts a suboptimal path characterized by numerous unnecessary steps being taken. The number of steps and the accumulated reward of figure 13.c are 33 and -18.5 respectively.

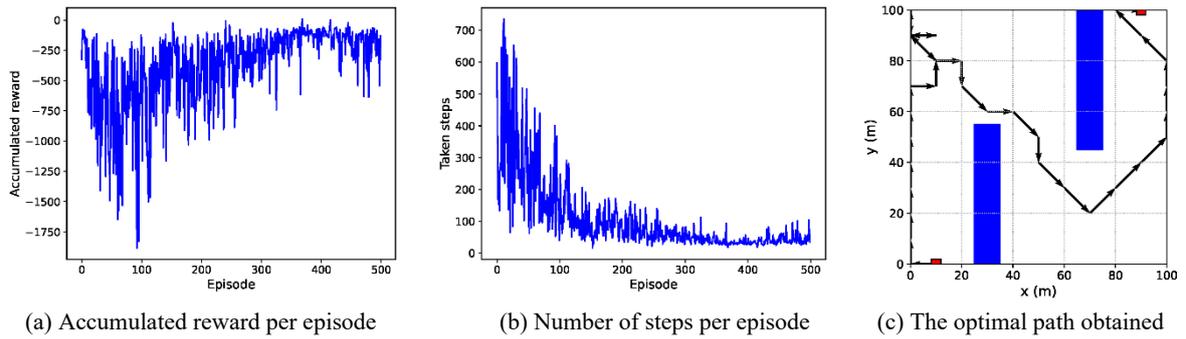

(a) Accumulated reward per episode  (b) Number of steps per episode  (c) The optimal path obtained

**Figure 15.** Experimental results of SARSA

# 4 Analysis and discussion

Section 3.1 introduces the simulation experiment of the improved ORB-SLAM3 system proposed in this project. In this section, this project has used the open-source TUM Dataset and the self-made dataset to test the Visual SLAM system. As shown in figure 6 and figure 10, the improved ORB-SLAM3 system successfully completed the dense point cloud mapping function using both the TUM dataset and the self-made dataset. The obtained dense point cloud maps are complete and can reflect the details of the real experimental environment. Analogous to the three parallel threads (tracking, local mapping, and loop closing) of ORB-SLAM3, the dense point cloud mapping thread is relatively independent and does not have an adverse impact on the real-time performance of the system. At the same time, a large amount of redundant map information is removed through the map conversion process shown in figure 4. The obtained grid maps from the experiments, which encompassed the essential information necessary for path planning, exhibited a storage footprint of merely 4.34KB and 3.75KB respectively. The back-end path planning can be carried out based on the obtained grid map.

Section 3.2 shows the experiments, which apply the three reinforcement learning algorithms: Q-learning, DQN, and SARSA to the same path planning problem. The performance of the three algorithms can be compared based on the experimental results shown in figure 13, 14, and 15. Among the three algorithms, the DQN algorithm converges the fastest and finds the optimal path from the starting point to the target point after 300 training episodes; followed by Q-learning, after 500 episodes of training, the accumulated reward per episode and the number of steps per episode all converge to stable values. But Q-learning ultimately failed to find the optimal path. The path shown in figure 13.c is close to the optimal path shown in figure 14.c. The SARSA algorithm has the worst performance. As shown in the figure 15, the accumulated reward per episode and the number of steps per episode fluctuate violently throughout the training process and fail to converge to stable values after 500 episodes of training. The SARSA algorithm ultimately did not find the optimal path, and there exists enormous difference between the path it found and the optimal path. Hyperparameters, except for the learning rates corresponding to different loss functions, were kept the same in the experiments. Experience replay is used in the training of Q-learning and DQN. This may be one reason why the convergence effect of the SARSA algorithm during the training process is so poor compared to the other two algorithms. Experience replay reduces the correlation of samples, which makes the training process more independent and stable. Besides, it allows experience samples to be used repeatedly, which improves the utilization rate of data, allowing the agent to learn more knowledge from past experiences. The difference in convergence speed and the final path planning effect between Q-learning and DQN reflects the advantages of deep neural networks in fitting value functions compared to the Q-table under the conditions of high-dimensional state space.

Based on the above analysis, it can be concluded that the improved ORB-SLAM3 system proposed in this project can meet the needs of dense point cloud mapping. The map conversion is the key step to connect the front-end Visual SLAM system and the back-end path planning, which makes the two independent parts into a complete system. In addition, this project has also compared three value learning algorithms and concluded that:

- The Q-learning algorithm using experience replay is more stable and converges faster than the SARSA algorithm.

- DQN has better performance in fitting the value function than Q-table when it comes to high-dimensional and complex problems.

However, there are still some drawbacks in this project that cannot be ignored, such as:

- This project only considers the global path planning. However, the local path planning requirements are more common in real-life robot applications

- The datasets used in the experiments are recorded by either controlling the robot to complete a loop in the environment or completing a collision-free path with the hand-held camera. The actual situation is that the robot has no information about the surrounding environment in advance. Collisions and inability to find the target point are inevitable during the dataset recording.

- This project treats the robot's actions as discrete to apply the three value learning algorithms. In real life, robot movements are often continuous and complex. The assumption of a discrete action space is not universal.

These drawbacks are not insurmountable and can be addressed in future work.

# 5 Conclusions and further work

## 5.1 Conclusions

With the rapid development of robotics technology, mobile robots are playing an increasingly important role in human life. The localization and autonomous navigation of robots in unknown environments is currently a research focus in the field of robotics. This project has conducted research on robot path planning based on Visual SLAM. The two key technologies: Visual SLAM and path planning has been optimized and improved. The experimental verifications have been conducted respectively. The main work of this project is as follows:

(1) **Construction of Visual SLAM system**. This project improves the open-source ORB-SLAM3 system and adds a dense point cloud mapping thread to it, enabling it to create dense point cloud maps containing rich information about the environment.

(2) **Path planning map construction**. Dense point cloud maps contain a large amount of redundant information that is unnecessary for path planning. This project removes redundant map information by converting the dense point cloud map into an octomap, and then converts the octomap into a grid map, which is the suitable map for two-dimensional path planning.

(3) **Research on path planning algorithm based on reinforcement learning**. This project has applied three reinforcement learning algorithms: Q-learning, SARSA, and DQN to global path planning scenarios. According to the results in Section 3.2.3 and the analysis in Section 3, DQN is the optimal algorithm suitable for high-dimensional environments.

(4) **Experimental verification.** This project has conducted experimental verifications of the proposed Visual SLAM system and the path planning algorithms based on reinforcement learning. In the verification experiment of the Visual SLAM system, this project has built an experimental environment and used an RGB-D camera to record the dataset in addition to directly using the open-source dataset.

This project has completed the construction of a complete autonomous robot navigation system from the construction of the Visual SLAM system to the research of path planning algorithms, achieving the original aims of the project. In the process of finishing the project, I have systematically studied the entire Visual SLAM framework and was able to make improvements based on the open-source project. In addition, I have also systematically studied the content of reinforcement learning and independently developed path planning algorithms based on reinforcement learning. The knowledge I have learned in this project will be helpful to my future graduate studies.

## 5.2 Suggestions for further work

As what has been mentioned in Section 4, there are still some drawbacks in this project. To achieve the aims of the project and achieve experimental results, many experimental conditions have been idealized. Future research work can focus on the following aspects:

(1) Consider an online scenario where Visual SLAM and path planning run simultaneously. Local path planning algorithms are suitable to this scenario because of the lack of the global map information.

(2) Apply RL algorithms suitable for continuous action space, such as PPO and DDPG, to path planning problems.

(3) Migrate the system from the simulation environment to a robot and test its performance in real experimental environment.

https://proceedings.neurips.cc/paper_files/paper/2003/hash/ee8fe9093fbbb687bef15a38facc44d2-Abstract.html

11. Stentz, 'Optimal and efficient path planning for partially-known environments', in Proceedings of the 1994 IEEE International Conference on Robotics and Automation, May 1994, pp. 3310–3317 vol.4. doi: 10.1109/ROBOT.1994.351061.

12. S. Koenig and M. Likhachev, 'Real-time adaptive A*', in Proceedings of the fifth international joint conference on Autonomous agents and multiagent systems, in AAMAS '06. New York, NY, USA: Association for Computing Machinery, May 2006, pp. 281–288. doi: 10.1145/1160633.1160682.

13. V. Bulitko and G. Lee, 'Learning in Real-Time Search: A Unifying Framework', Journal of Artificial Intelligence Research, vol. 25, pp. 119–157, Feb. 2006, doi: 10.1613/jair.1789.

14. S. Koenig and M. Likhachev, 'D*lite', in Eighteenth national conference on Artificial intelligence, USA: American Association for Artificial Intelligence, Jul. 2002, pp. 476–483.

15. Stentz, 'The Focussed D* Algorithm for Real-Time Replanning'.

16. S. LaValle, 'Rapidly-exploring random trees : a new tool for path planning', The annual research report, 1998, Accessed: Jan. 24, 2024. [Online]. Available: https://www.semanticscholar.org/paper/Rapidly-exploring-random-trees-%3A-a-new-tool-for-LaValle/d967d9550f831a8b3f5cb00f8835a4c866da60ad

17. J. J. Kuffner and S. M. LaValle, 'RRT-connect: An efficient approach to single-query path planning', in Proceedings 2000 ICRA. Millennium Conference. IEEE International Conference on Robotics and Automation. Symposia Proceedings (Cat. No.00CH37065), Apr. 2000, pp. 995–1001 vol.2. doi: 10.1109/ROBOT.2000.844730.

18. S. Karaman and E. Frazzoli, 'Sampling-based Algorithms for Optimal Motion Planning'. arXiv, May 05, 2011. Accessed: Jan. 24, 2024. [Online]. Available: http://arxiv.org/abs/1105.1186

19. J. Bruce and M. Veloso, 'Real-time randomized path planning for robot navigation', in IEEE/RSJ International Conference on Intelligent Robots and Systems, Sep. 2002, pp. 2383–2388 vol.3. doi: 10.1109/IRDS.2002.1041624.
– 41 –

# B     University's Plagiarism Statement

This statement is reviewed annually and the definitive statement is in the [University Calendar](University Calendar).

The University's degrees and other academic awards are given in recognition of a student's personal achievement. All work submitted by students for assessment is accepted on the understanding that it is the student's own effort.

Plagiarism is defined as the submission or presentation of work, in any form, which is not one's own, without acknowledgement of the sources. Plagiarism includes inappropriate collaboration with others. Special cases of plagiarism can arise from a student using his or her own previous work (termed auto-plagiarism or self-plagiarism). Auto plagiarism includes using work that has already been submitted for assessment at this University or for any other academic award.

The incorporation of material without formal and proper acknowledgement (even with no deliberate intent to cheat) can constitute plagiarism. Work may be considered to be plagiarised if it consists of:

a direct quotation;

a close paraphrase;

an unacknowledged summary of a source;

direct copying or transcription.

With regard to essays, reports and dissertations, the rule is: if information or ideas are obtained from any source, that source must be acknowledged according to the appropriate convention in that discipline; and any direct quotation must be placed in quotation marks and the source cited immediately. Any failure to acknowledge adequately or to cite properly other sources in submitted work is plagiarism. Under examination conditions, material learnt by rote or close paraphrase will be expected to follow the usual rules of reference citation otherwise it will be considered as plagiarism. Departments should provide guidance on other appropriate use of references in examination conditions.

Plagiarism is considered to be an act of fraudulence and an offence against University discipline. Alleged plagiarism, at whatever stage of a student's studies, whether before or after graduation, will be investigated and dealt with appropriately by the University.

The University reserves the right to use plagiarism detection systems, which may be externally based, in the interests of improving academic standards when assessing student work.